\documentclass{article} % For LaTeX2e
\usepackage{iclr2026_conference,times}

\usepackage{amsmath,amsfonts,bm}

\def\eqref#1{equation~\ref{#1}}
\def\1{\bm{1}}

\DeclareMathAlphabet{\mathsfit}{\encodingdefault}{\sfdefault}{m}{sl}
\SetMathAlphabet{\mathsfit}{bold}{\encodingdefault}{\sfdefault}{bx}{n}

\usepackage{hyperref}
\usepackage{url}
\usepackage{graphicx}
\usepackage{amsmath,amssymb}
\usepackage{booktabs}
\usepackage{cleveref}
\usepackage{tcolorbox}
\usepackage{fontenc}

\title{A transformer architecture alteration to incentivise externalised reasoning}

\author{
Elizabeth Pavlova$^{1*}$ \quad
Mariia Koroliuk$^{1*}$ \quad
Karthik Viswanathan$^{1,2*}$ \\[5pt]
{\bf Cameron Tice}$^{\dagger}$ \quad
{\bf Edward James Young}$^{\dagger}$ \quad
{\bf Puria Radmard}$^{\dagger}$ \quad
\\[5pt]
{\normalsize $^1$MARS} \\
{\normalsize $^2$University of Amsterdam} \\
{\normalsize $^*$Equal contribution} \\
{\normalsize $^\dagger$Geodesic Research, equal contribution} \\
\\ %[3pt]
\texttt{lizapavlova565@gmail.com,maria.koroliuk@gmail.com}
}

\iclrfinalcopy % Uncomment for camera-ready version, but NOT for submission.
\begin{document}

\maketitle

\begin{abstract}
We propose a new architectural change, and post-training pipeline, for making LLMs more verbose reasoners by teaching a model to truncate forward passes early. We augment an existing transformer architecture with an early-exit mechanism at intermediate layers and train the model to exit at shallower layers when the next token can be predicted without deep computation. After a calibration stage, we incentivise the model to exit as early as possible while maintaining task performance using reinforcement learning. We provide preliminary results to this effect for small reasoning models, showing that they learn to adaptively reduce computations across tokens. We predict that, applied at the right scale, our approach can minimise the amount of excess computation that reasoning models have at their disposal to perform non-myopic planning using their internal activations, reserving this only for difficult-to-predict tokens. 
%
%
% By reducing the computation available within hidden layers for easy-to-predict tokens,
%we expect this technique to improve monitorability by forcing 
% we train models to externalise an increased proportion of their reasoning into chain-of-thought tokens rather than performing it internally.
%\textcolor{blue}{we train models to externalize reasoning that would have been [located] in a single forward pass into additional chain-of-thought tokens for similar computational costs.}
%
% We implement our adaptive early exit architecture on \textcolor{red}{DeepSeek-R1-Distill-Qwen-1.5B} and present preliminary results demonstrating that the mechanism behaves as intended: the model learns to adaptively vary computational depth per token, achieving up to \textcolor{red}{22.4\% average layer} savings with a tunable trade-off against generation quality. Our approach introduces a tunable parameter that allows developers to modulate the strength of externalisation pressure based on specific task requirements, creating a negative safety tax by simultaneously reducing computational costs and improving oversight.
\end{abstract}

\begin{figure}[h!]
    \begin{center}
    \includegraphics[width=\linewidth]{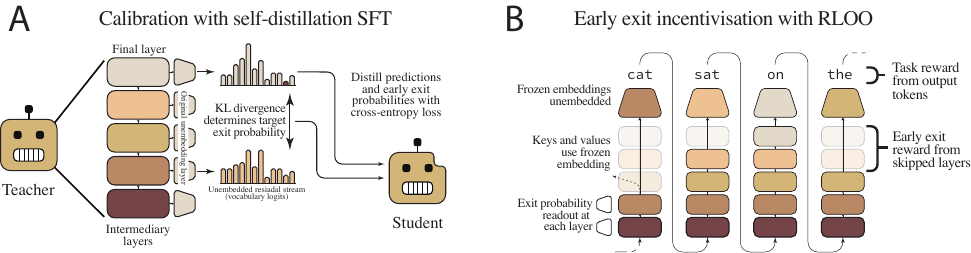}
    \end{center}
    \caption{\textbf{Overview of the early exit architecture.} \textbf{(A)} Calibration via self-distillation: early-exit heads at intermediate layers sample exit decisions from learned probability distributions. Exit probabilities are calibrated against a teacher (the full-depth model) where KL divergence between intermediate and final layer logits determines targets, trained alongside token predictions (Appendix \ref{app:sft}) \textbf{(B)} Early exit incentivisation via RL: the model generates sequences using the learned exit mechanism, with task reward from output tokens and early exit reward from skipped layers. Layers above the exit point are skipped, and the frozen residual stream representation is passed directly to the unembedding layer.}
    \label{fig:architecture}
\end{figure}

\section{Introduction}

% Given that this is for a general audience rather than specifically an AI safety audience (i.e., ICLR rather than LW), I would include a paragraph here about the general use of CoT---What it is, how it's trained, what kind of tasks it's used for. Doesn't have to be long but I think it's necessary for establishing background

Chain-of-Thought (CoT) reasoning in Language Models allows models to produce a series of intermediate ``reasoning'' tokens before arriving at a final answer \cite{wei2023chainofthoughtpromptingelicitsreasoning}. Although CoT began as a prompting strategy, it is now standard practice to train models to produce CoT via reinforcement learning \cite{openai2024openaio1card,DeepSeekAI2025}. Models often struggle to perform complex tasks without using their CoT to perform serial reasoning steps steps~\citep{emmons2025chainthoughtnecessarylanguage,arnav2025cotredhandedstresstesting}. Accordingly, many AI safety researchers are converging around the idea that the CoT of models may be a valuable tool for monitoring the actions they may take before they do so \citep{Korbak2025}.

%There are a number of imminent threats to the use of CoT monitoring for control:
However, the use of CoT monitoring is threatened by:
(i) \textbf{unfaithfulness}, where CoT fails to reflect the actual cognition of the model~\citep{Turpin2023,lanham2023measuringfaithfulnesschainofthoughtreasoning}, including malicious intent or planning; and
(ii) \textbf{obfuscation}, where optimisation pressure placed on CoT renders undesired reasoning invisible to a monitor without preventing the reasoning itself~\citep{Baker2025, Greenblatt2025, drori2025outputsupervisionobfuscatechain};

Obfuscated reasoning is facilitated by the fact that transformers are \emph{non-myopic}: each forward pass is optimised not just for the immediate next token, but for subsequent predictions, with models shown to pre-cache information for tokens generated several positions later~\citep{Pal2023, lindsey2025biology, Wu2024, Janus2023}. When the next token is easy to predict, training pressure incentivises using remaining compute for hidden computations useful later on---in the extreme, models can be trained to solve problems using only internal compute, without any externalised serial reasoning steps~\citep{pfau2024letsthinkdotdot, greenblatt_recent_2025}.

This work provides a proof of concept for an \textbf{architectural change that incentivises externalisation}, and a two-stage training method (self-distillation followed by reinforcement learning) where the model is rewarded for giving up as much of this hidden compute as possible, while still performing well. We present preliminary RL results showing models can exit earlier while maintaining task performance. We leave the evaluation of faithfulness for future work.

%The remainder of this paper is structured as follows. First, we outline why a naive proposal for increasing externalisation -- using a small model with extensive reasoning training -- is unlikely to succeed. By examining where this proposal fails, we arrive at a new direction: externalisation via adaptive compute. We give a concrete implementation of our adaptive compute architecture, along with a training pipeline aimed at increasing externalisation of model cognition when possible. We provide preliminary results from this architecture and highlight some interesting observed phenomena we hope to explore further as we continue this research.

Prior work has shown that unembedding intermediate layer activations results in coherent next token predictions that converge to final layer predictions~\citep{nostalgebraist2020}. Trained early-exit classifiers like ours can identify layers within a model's forward pass where early exits are possible without changing the final output~\citep{Schuster2022}.
We differ from prior work in two key ways. % Mechanically, when a model (stochastically) exits early at layer $k$, it stops writing to its residual stream and passes the frozen representation forward to the final unembedding layer to predict the next token. 
First, our mechanism is stochastic rather than deterministic, as at each layer the model samples from a probability distribution determining whether to exit. This enables our second novel contribution in training using reinforcement learning to incentivise even earlier exits.

\section{Methods}

\label{sec:architecture}

We make a simple architectural change that allow the transformer to stop writing to its residual stream at intermediary layers. We add shallow early-exit heads (with one early exit head every 5 layers), from which we read out the probability of early exit at a layer. In addition to the early-exit heads, we train LoRA adapters attached to the existing model weights \citep{Hu2022}.
The early-exit heads project the embeddings of exitable layers to a scalar logit. During generation, the model will stochastically `exit' at that layer according to this logit.
% which is converted to a probability of exiting at that layer during inference time forward passes. 
When the model exits at a layer, the residual stream passes unchanged through the remaining layers, to the unembedding layer.
Keys and values from later layers are derived from this frozen residual stream (Figure \ref{fig:architecture}B). 
%
% The logit can also be offset by a selected amount, altering the propensity to exit early. This offset can be tuned according to the context/model requirements. We will return to this design aspect shortly.
%
% This approach builds upon established prior findings. 
%
We train the learnable weights of our architecture (LoRA adapters and exit probes) using a two-stage training procedure: supervised fine-tuning followed by reinforcement learning. During the latter, we train the model to reduce the number of layers used on average, rather than just identifying when exits are safe.

\textbf {Training Stage 1: Early exit calibration through self-distillation} \quad
The modified model is trained to match the original model's full-depth predictive logits, a technique we call \emph{self-distillation}. We treat the original full-depth model as a frozen teacher and train the early-exit variant as the student, augmented with the new parameters. % The model is allowed to ``exit early'', by freezing the residual stream at layer $k$ and passing it unchanged through all remaining layers until the final unembedding.
We use an information theoretic heuristic to determine which layers \textit{would have been} reasonable for the original model to exit from, and train the student on a combination of token distribution similarity (as in normal distillation), and recovery of this exit likelihood at each layer (Figure \ref{fig:architecture}A).
This heuristic---alongside the complete SFT training algorithm---is provided in Appendix~\ref{app:sft}. There, we also empirically verify that early exiting with probabilities defined by this heuristic target results in coherent sentences, and hence that this is a reasonable target of replication for our student model (Appendix~\ref{app:heur_ver}).

\textbf{Training Stage 2: Reinforcement learning to incentivise earlier exits} \quad
Next, we apply reinforcement learning with an explicit reward for exiting earlier (reducing depth used per-token) and a penalty for task failures. While stage 1 teaches the model when it can exit early without changing behaviour, stage 2 actively incentivises exiting as early as possible.

We develop a modified version of RLOO~\citep{ahmadian2024basicsrevisitingreinforcestyle} with an additional reward term that penalises the average layer depth at which exits occur:
\begin{equation}
    R_{\text{total}} = R_{\text{task}} - \lambda \cdot (\text{average exit layer index}) - \beta \cdot D_{\text{KL}}(\text{policy} \,\|\, \text{base model}),
\end{equation}
where
$R_{\text{task}}$ is the standard task reward (\emph{e.g.}, correctly answering questions); the second term penalises using later layers, rewarding earlier exits; and the final term is a  regulariser over the output vocabulary of the policy.

The hyperparameter $\lambda$ effectively controls the ``eagerness'' of early exits rewarded during training - how aggressively we are willing to trade off accurate task performance for reduced internal computation. When high, it penalises the average exit layer, ensuring the model does better if it produces more tokens with earlier average exits per token.
This gives us a tunable dial that safety-critical applications can dial up to maximise reasoning externalisation, while performance-critical applications can reduce it to ensure task completion. The $\beta$ term acts as a standard regulariser to prevent the model from drifting too far from its original behaviour.
%
% We provide more commentary on our method in Appendix \ref{app:sft}, describing how our modified version of RLOO can be seen as normal transformer RL in an augmented vocabulary that also covers the exit layer index.
%
% \textcolor{red}{IF TIME/COGNITIVE BANDWIDTH - FULL RLOO MATHS WOULD BE GREAT IN APPENDIX... EDWARD?}
%
A similar dial is afforded at inference time in the form of early exit logit scaling. We do not use this feature, but discuss its implications in Section~\ref{sec:discussion}.

Probabilistic quantities required in RLOO, such as perplexities of sampled sequences, also incorporate the probability of early exits. This is equivalent to an augmentation of the action space/vocabulary to include binary exist decisions \textit{at each layer} (Appendix \ref{app:sft}).

%\cam{I'm slightly confused about the lambda parameter. Here we're implying that you can set the parameter to change RL training, but my understanding of this variable is that we could use it to change the eagerness of ealy exiting at inference time?}

\section{Results}
\label{sec:results}

We implemented our adaptive early exit architecture on two smaller reasoning models, and applied the self-distillation SFT on both GSM8K~\citep{Cobbe2021} and a theory of mind dataset generated with the open-source ExploreToM implementation~\citep{Sclar2024}. Below are our preliminary findings demonstrating that the early exit mechanism behaves as intended. However, we stop short of verifying the intended effects on CoT monitorability.

\begin{figure*}
\begin{center}
\includegraphics[width=\linewidth]{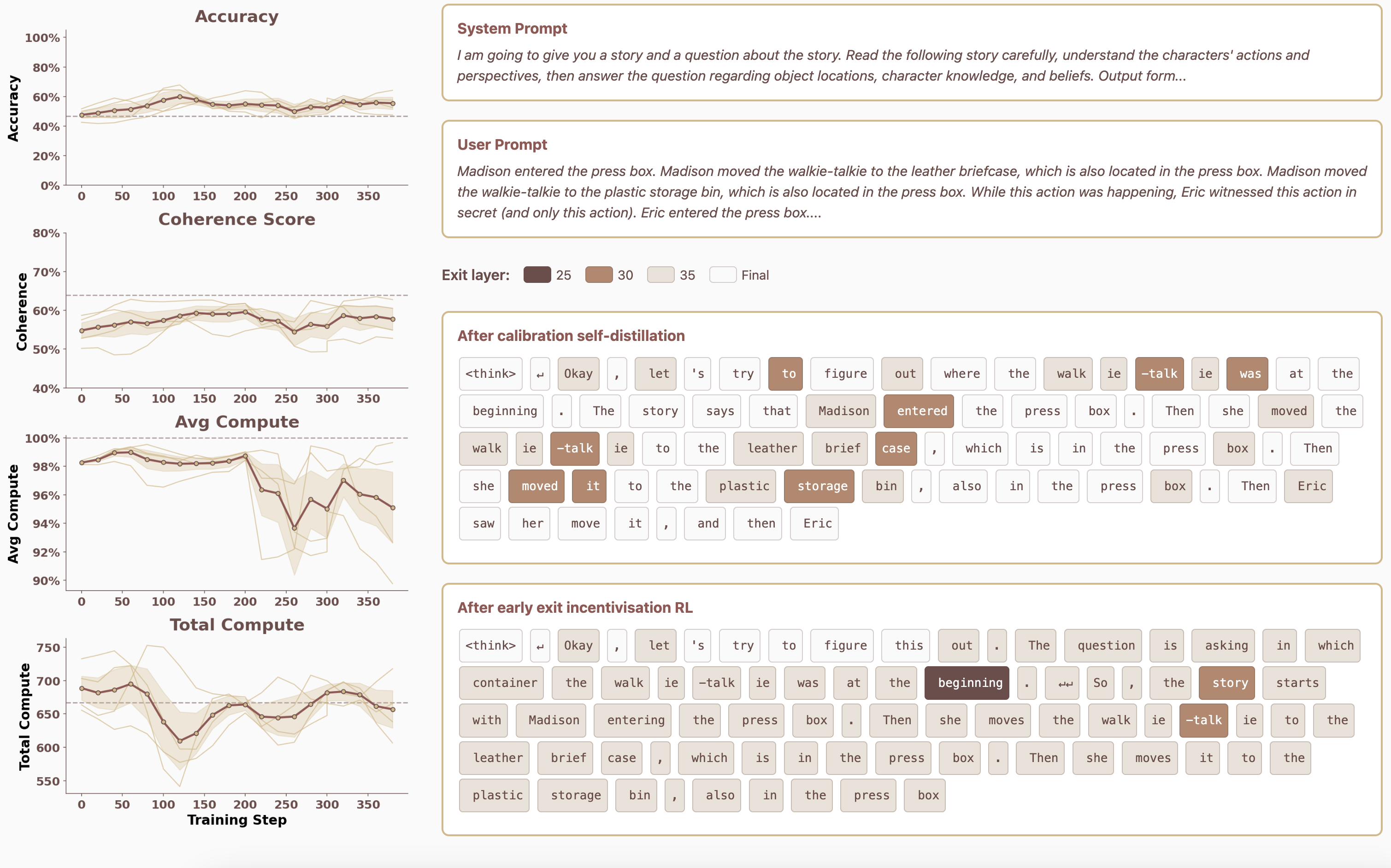}
\end{center}
\vspace{-2em}
\caption{\textbf{Left:} RL training dynamics on Qwen3-4B for the Theory of Mind \citep{Sclar2024} task. Each panel shows individual runs (light lines), mean across runs (bold line with markers), and $\pm$1 SE error bars. The dashed line indicates the pre-SFT base model performance. Step 0 represents the SFT-calibrated model before RL. Accuracy improves while average compute decreases, demonstrating that the model learns to exit earlier without sacrificing performance. Coherence remains stable throughout. Evaluated on 60 theory of mind prompts per step. \textbf{Right:} Token-level visualisation of early exit behaviour. Each token is coloured by the layer at which computation was terminated. The model adaptively varies computational depth, using fewer layers for predictable tokens and the full depth for more complex ones. We used $\lambda=1.5$ and $\beta=0.25$.}
\vspace{-1em}
\label{fig:main_results}
\end{figure*}

\paragraph{Self-distilled early exits match target distribution}
During Stage 1 training, we constructed target exit probabilities by computing KL divergence between each layer's predictions and the final layer outputs. For each token, an exit layer is sampled from this distribution and the model is trained to match these teacher targets using cross-entropy loss.
In the interest of space, we defer all results to Appendix \ref{app:dist_ver}, where we show that the for DeepSeek-R1-Distill-Qwen-1.5B~\citep{DeepSeekAI2025}, we were able to \textit{calibrate} the early exit parameters to achieve the same rate of early exits as in the self-distillation training data.
This provides a good starting point for the RL phase.

\textbf{Reinforcement learning decreases forward pass compute while maintaining coherence and accuracy.} \quad
Having demonstrated that our adaptive early exit mechanism works as intended on DeepSeek-R1-Distill-Qwen-1.5B, we now evaluate whether reinforcement learning can incentivise earlier exiting while maintaining output quality. For these experiments, we use Qwen3-4B, applying the same Stage 1 SFT calibration procedure before RL training.
%
%\textcolor{red}{[VERIFY THESE NUMBERS - at one point there were hyperparmeters here too but I can't find them.]} lambda and beta in caption.
%
Figure~\ref{fig:main_results} (left) shows RL training dynamics across five runs on Qwen3-4B after SFT calibration, evaluated on 60 prompts per checkpoint. We track four metrics: task accuracy, answer coherence, average compute (proportion of total layers used per token), and total compute (CoT length in tokens $\times$ average compute). Task accuracy improves over the RL stage of training, rising from 47\% at step 0 (directly after self-distillation) to 55-60\% by step 380, consistently exceeding the pre-SFT base model (dashed line). \textbf{Average layer compute decreases from 98\% to approximately 95\% with some runs dropping to a 90\% average, indicating the model learns to exit earlier.} 

% Total compute decreases despite increases in chain-of-thought length, indicating that the per-token compute savings from early exiting offset any increases in generation length. 
Answer coherence is stable across RL training but sits slightly below the unmodified baseline. The variance across runs (light lines) shows some instability in later training steps, though the mean trajectory (bold line) demonstrates consistent improvement.
The model adopts a policy of consistent last layer removal, with some easier tokens, such as the `\texttt{-talk}' in `walkie-talkie' requiring even less internal compute.
\vspace{-0.5em}

%below chart will be updated with additional panel on right side showing pre RL rollout on top and post on bottom

\section{Discussion}
\label{sec:discussion}

\vspace{-0.5em}
These findings provide a proof of concept that our adaptive early exit architecture works as intended. 
% Our SFT experiments on DeepSeek-R1-Distill-Qwen-1.5B demonstrate that the model learns the intended exit behaviour, with exit distributions closely matching training targets. 
Our RL results on Qwen3-4B show stable coherence and improving accuracy alongside slight decreases in average computation, suggesting that the model can learn to exit earlier while actually enhancing task performance.
While our preliminary results are promising, several limitations and concerns remain:

% REINTRODUCE IF NEEDED
% \paragraph{Faithfulness vs.\ rationalisation.} Once we finalise our RL pipeline to consistently incentivise earlier exits, we must address a core assumption of this work: forcing early exits drives genuine reasoning into the token stream. However, there is a risk that the model simply learns to output ``filler'' tokens that mimic reasoning to buy computational time, effectively post-hoc rationalising a decision made in the initial layers. We need to rigorously test whether the externalised CoT is causal to the final answer by conducting intervention experiments, such as CoT token replacement, to verify that the generated reasoning is truly load-bearing.

\textbf{Extension to different domains.}\quad Our preliminary evaluations have been on within-task performance for GSM8K and theory of mind data and we still need to run cross-domain validation to evaluate generalisation of our early-exit architecture. Additionally, validating our method with more complex datasets on larger models will be beneficial in future work.

%\paragraph{Further scaling.} Our SFT experiments were conducted on a 1.5B parameter model, with RL experiments scaling to 4B parameters. It remains an open question whether larger models and different model families will be as responsive to this architecture. %Validating this is essential before it can inform safety practices and is a primary objective for us once additional compute budget becomes available.

\textbf{Mitigating quality trade-offs.}\quad As our training pushes exits to even earlier layers, care is needed to ensure overall model performance does not drop. Our method introduces new hyperparameters such as the KL factor during self-distillation SFT and the $\lambda$ weighting parameter in RL designed to balance externalisation with task accuracy. Our current RL results suggest this balance is achievable, though further hyperparameter tuning can result in additional improvements. For instance, inference time interventions such as changing the threshold at which early exiting occurs could be used to trade-off between safety (through CoT vebosity and monitorability), and original model quality in less safety-critical domains.

\section{Conclusion}
\label{sec:conclusion}

We have presented a post-training method for increasing the externalisation of reasoning in large language models through adaptive early-exit mechanisms. 
%
%Our approach augments existing architectures with learnable exit heads that allow the model to terminate computation at intermediate layers when the next token can be predicted with shallow processing. 
%
Through a two-stage training pipeline consisting of supervised calibration followed by reinforcement learning, we incentivise models to use less compute per token, pushing complex reasoning into human-understandable chain-of-thought.
Our preliminary experiments demonstrate that the early exit mechanism works as intended: the model learns to adaptively vary computational depth per token, with a tunable trade-off between externalisation pressure and generation quality. We are excited for this concept to be scaled from a proof of concept to a rigorous safety tool and believe this direction represents a promising and neglected avenue for improving the monitorability of frontier models.

%welcome collaborators to help scale this work 
\clearpage

% REINTRODUCE AFTER ACCEPTANCE
% \subsubsection*{Acknowledgments}
% This work was done in Mentorship for Alignment Research Students (MARS) 3.0 in a Geodesic Research stream. We would like to thank the Cambridge AI Safety Hub (CAISH) for organizing the MARS program, and Meridian Cambridge for providing the community and venue for collaboration.

\bibliography{iclr2026_conference}
\bibliographystyle{iclr2026_conference}

\clearpage
\appendix

\section{Weight calibration with SFT - more details}
\label{app:sft}
Our architecture can be viewed as performing next-token prediction over an augmented vocabulary of layer-token pairs $(\ell, v) \in L \times V$. Instead of only predicting the next token $v$, the model implicitly predicts a joint event---exit at layer $\ell$ and emit token $v$---with joint probability
\begin{equation}\label{eq:augmented_vocabulary}
    p(\ell, v) = p(\ell) \, p(v \mid \ell),
\end{equation}
where the early exit decision heads determine $p(\ell)$, and $p(v \mid \ell)$ is obtained by unembedding the residual stream at layer $\ell$ using the model's unembedding matrix.

\textbf{This framing lets us treat early exit as a standard next-token prediction problem over an enlarged vocabulary}, so that techniques developed for standard next-token prediction (such as supervised fine-tuning and RLVR) carry over directly to this paradigm.

We create the training data for the weight calibration SFT stage by running the base model on a dataset and record two things at each layer: the next-token probability distribution (computed by applying the unembedding matrix to intermediate layer's activations), and how similar this distribution is to the final layer's prediction (measured by forward KL divergence). When an intermediate layer's predictions are very close to the final layer (low KL divergence), we know that the token could have been predicted early.
This heuristic method for deriving training data for the early exit heads is depicted in Figure \ref{fig:architecture}A.

By passing these KL divergence values into a temperature-modulated softmax function over layers, we construct a target exit probability distribution where layers with lower divergence are assigned higher exit probabilities. Specifically, we transform KL divergences into a probability distribution over layers using:
\begin{equation}\label{eq:early_exit_teaching_probs}
    p(\ell = k) = s_k \cdot \prod_{j=0}^{k-1} (1 - s_j),
\end{equation}
where
\begin{equation}\label{eq:early_exit_layer_probs}
    s_k = 2\left(1 - \sigma\!\left(\text{KL}_{\text{factor}} \cdot D_{\text{KL}}^{(k)}\right)\right),
\end{equation}
and $D_{\text{KL}}^{(k)}$ is the KL divergence between layer $k$'s predictions and the final layer. This ensures that layers with lower KL divergence (more similar to final output) are assigned higher exit probabilities. The final layer is a catch-all for any remaining probability mass if no early exit occurs. The KL factor controls the temperature of this transformation, with higher values requiring greater confidence (lower KL divergence) before assigning high exit probabilities. Combined with the teacher model's next-token distribution, this gives us target joint probabilities $p(\ell, v)$ for training.

The loss function consists of two terms: a match on the original model's next-token predictions (KL divergence between the trained model's output and the teacher model's final layer output) and a match on the exit probability distribution (cross entropy loss on sampled exit layer):
\begin{equation}\label{eq:sft_loss_function}
    \mathcal{L} = D_{\text{KL}}\!\left[p_{\text{student}}(y) \,\|\, p_{\text{teacher}}(y)\right] - \log p_{\text{model}}(e_{\ell})
\end{equation}
where $y$ is the next token being predicted. Essentially, the model acts as its own teacher, having learned when more limited computation suffices to replicate what the full forward pass would produce.

Once trained, the model samples from the exit probabilities during its forward pass at inference time, creating a stochastic early-exit process. 
While we never pull this lever, the exit logit can also be offset by a selected amount, altering the propensity to exit early. Positive offsets increase the probability of early exits (favouring externalisation), while negative offsets decrease it (allowing for more internal computation).
This essentially allows an inference-time counterpart to $\lambda$ as a controller on the eagerness of exits.

As we demonstrate in our experiments below, we have already observed this behaviour at $\sim$1.5B scale: on many tokens, the model uses fewer layers while reproducing its own reasoning trajectories.

\subsection{Target distribution verification}
\label{app:heur_ver}

Using DeepSeek-R1-Distill-Qwen-1.5B, we now verify that this KL-based heuristic is a reasonable one for the student model to replicate.
The KL factor is a temperature parameter in the sigmoid function that controls the exit distribution during SFT training---see \cref{eq:early_exit_layer_probs}. It modulates how the model converts KL divergence values (similarity between intermediate and final layer predictions) into exit probabilities. We trained models at five settings (KL factors: 0.25, 0.5, 1.0, 2.0, 4.0), where lower values encourage earlier exits and higher values need greater confidence, measured through lower KL divergence.

To evaluate the effect of this parameter, we tested the same 50 GSM8K prompts across the base model and all five early-exiting models. To assess whether the early-exit mechanism affects reasoning quality, we evaluated outputs using GPT-5 as an automated judge scoring generations across four qualitative dimensions: coherence, completeness, clarity, and repetition avoidance. See Appendix~\ref{app:scorer} for the scorer prompt.

\begin{figure}[t]
    \begin{center}
    \includegraphics[width=0.8\linewidth]{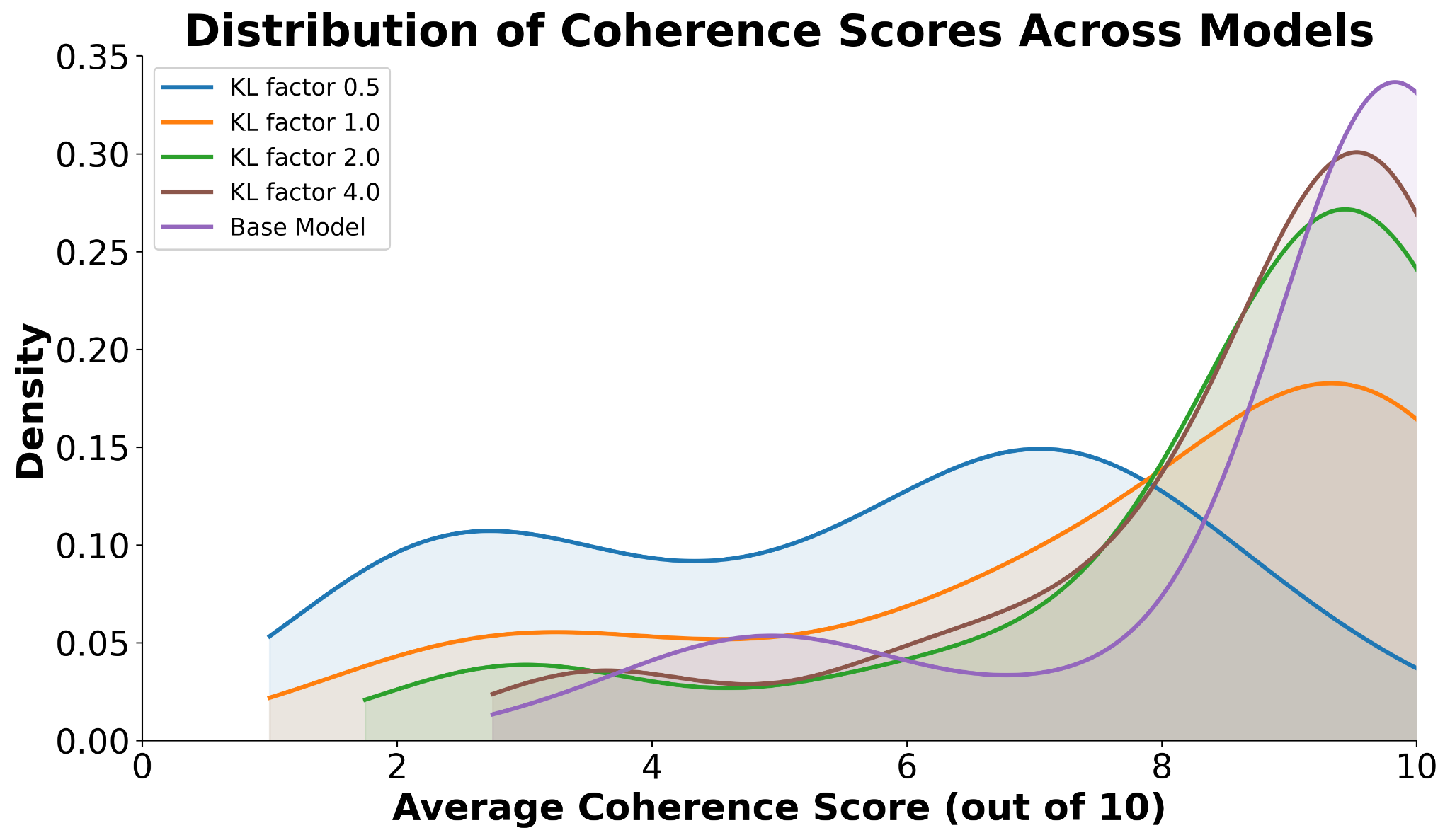}
    \end{center}
    \caption{\textbf{Distribution of coherence scores across models with different KL factors.}\protect\footnotemark[3] As the KL factor increases, coherence distributions shift toward higher scores, approaching base model performance.}
    \label{fig:coherence}
\end{figure}

\footnotetext[3]{Model trained on KL factor=0.25 is excluded from this visualisation due to its outlier behaviour (scores were heavily concentrated around 1, compressing the scale for other models) but is included in Table~\ref{tab:results}.}

To further quantify the trade-off between quality and computational efficiency, we tracked Average Layer Savings (ALS), which measures the reduction in computational depth averaged across all generated tokens:
\begin{equation}
    \text{ALS}(\%) = \frac{1}{N} \sum_{i=1}^{N} \left(1 - \frac{\ell_{\text{exit}}(i)}{L_{\text{total}}}\right)
\end{equation}
where $N$ is the sequence length, $\ell_{\text{exit}}(i)$ is the layer index where the exit occurred, and $L_{\text{total}}$ is the total layer count (28 for DeepSeek-R1-Distill-Qwen-1.5B).

\begin{table}[t]
\caption{\textbf{Effect of KL factor on coherence, early exit rate, and average layer savings.} Aggressive early exits (KL factor=0.25) achieve the highest computational savings but at a large cost to coherence. KL factor=1.0 balances reasonable computational savings with strong coherence.}
\label{tab:results}
\begin{center}
\begin{tabular}{lccc}
\toprule
\textbf{Model} & \textbf{Avg. Coherence} & \textbf{Early Exit Rate} & \textbf{Avg. Layer Savings} \\
\midrule
KL factor 0.25 & 1.1 & 59\% & 22.4\% \\
KL factor 0.5  & 5.5 & 49\% & 8.1\% \\
KL factor 1.0  & 7.5 & 47\% & 6.4\% \\
KL factor 2.0  & 8.4 & 43\% & 5.3\% \\
KL factor 4.0  & 8.5 & 37\% & 4.6\% \\
Base Model     & 8.9 & 0\%  & 0.0\% \\
\bottomrule
\end{tabular}
\end{center}
\end{table}

Based on this analysis, we selected KL factor=1.0 as our standard setting for subsequent experiments, as it offered a balance between externalisation pressure and task performance.

\subsection{Distillation training verification}
\label{app:dist_ver}

\begin{figure}[t]
\begin{center}
\includegraphics[width=0.8\linewidth]{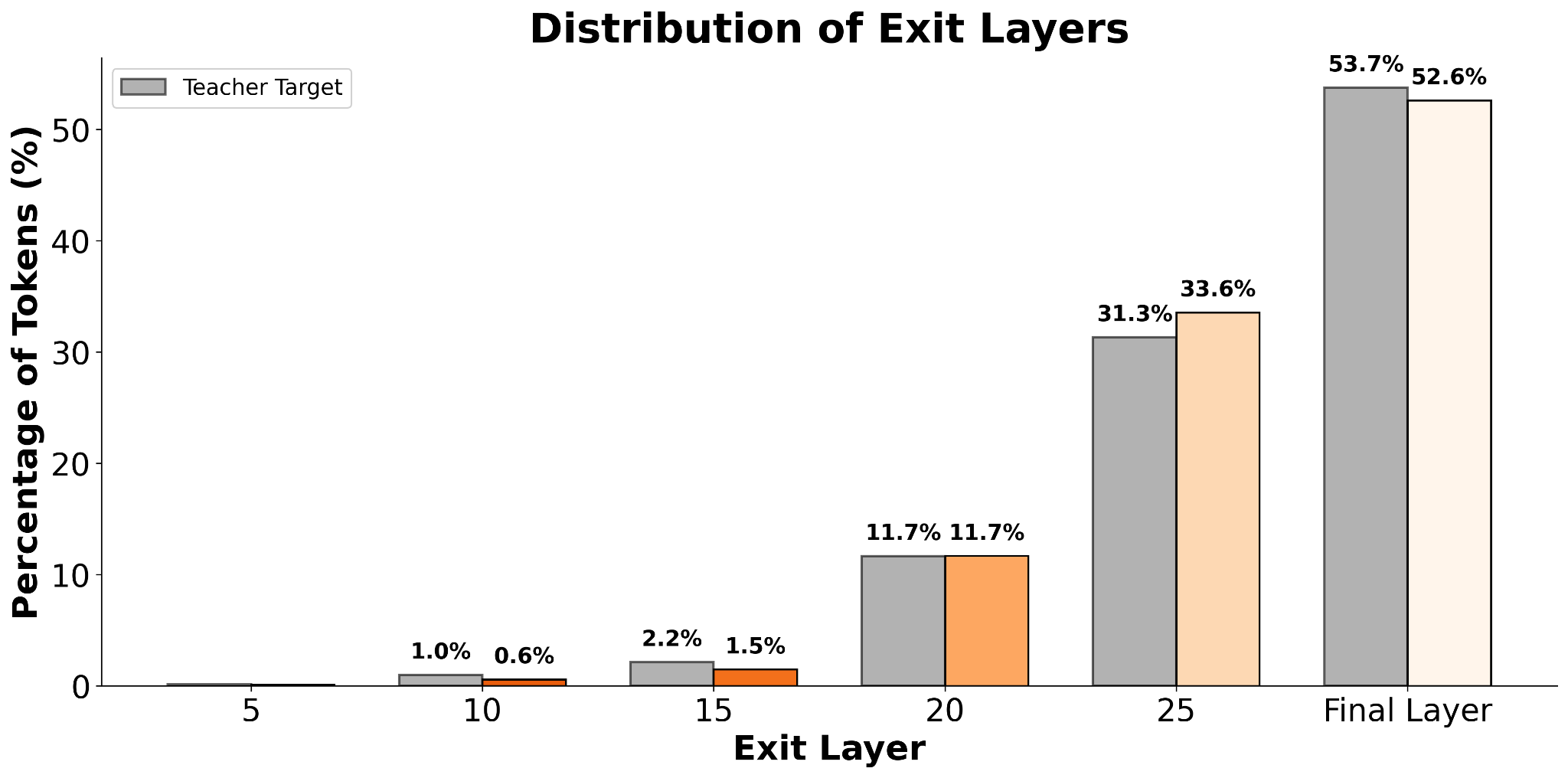}
\end{center}
\caption{\textbf{Distribution of exit layers.} The model's learned exit distribution (orange bars) closely follows the target exit probability distribution from training data (grey bars), demonstrating effective learning of exit behaviour during Stage 1 calibration.}
\label{fig:exitdist}
\end{figure}

To evaluate whether our trained model learned these exit patterns, we tested on 50 GSM8K questions. Figure~\ref{fig:exitdist} shows an overall 47\% early exit rate on tokens from the 50 responses, concentrated in later layers (20--25). The model's learned exit distribution closely follows the target exit probability distribution from our training data, showing effective learning of the model's intended exit behaviour during Stage 1 calibration.

\clearpage
\section{Scorer Prompt}
\label{app:scorer}

For scoring, we use the following prompt:

\begin{tcolorbox}[title=Scorer prompt, colback=white, colframe=darkgray]
\begin{small}
\small
You are evaluating the coherence and quality of an AI assistant's chain of thought and response. Allow exploratory/meandering reasoning as long as it stays internally consistent.

Original Question: \{prompt\}

AI Response: \{response\}

Please evaluate this response on the following criteria:
\begin{enumerate}
    \item Coherence and logical flow (1--10 scale): Does the reasoning follow a sensible progression, even if unfinished?
    \item Completeness of the reasoning (1--10 scale): Does it reach a correct and explicit chain of thought? If partial but on-track and relevant to answer, award mid-range (4--7).
    \item Clarity and readability (1--10 scale): Is it easy to follow? Minor meandering is okay if readable.
    \item Absence of repetition or errors (1--10 scale): Penalise contradictions, factual mistakes about the prompt, or heavy repetition.
\end{enumerate}

Rate each criterion and provide an overall score from 1--10:
1: major breakdown (nonsensical, off-topic);
4: noticeable issues but some useful reasoning;
7: generally solid, with minor flaws or cut-offs;
10: excellent, complete, and polished.

Format your response as:\\
Coherence: X/10\\
Completeness: X/10\\
Clarity: X/10\\
No Repetition: X/10\\
Overall: X/40\\
Brief explanation: [your reasoning]
\end{small}
\end{tcolorbox}

\end{document}